\documentclass[letterpaper, 10 pt, journal, twoside]{IEEEtran}  

\IEEEoverridecommandlockouts                              

\usepackage{graphics} 
\usepackage{epsfig} 
\usepackage{amsmath} 
\usepackage{amssymb}  
\usepackage{hyperref}
\usepackage{xcolor}
\usepackage{tikz}
\usetikzlibrary{calc}
\usetikzlibrary{arrows}
\usepackage{pgfplots}
\usepackage{siunitx}
\newcommand{\argmin}{\operatornamewithlimits{arg\ min}}

\pgfplotsset{compat=1.18}

\title{Geo-localization based on Dynamically \\ Weighted Factor-graph
}

\author{Miguel Ángel Muñoz-Bañón,  Alejandro Olivas,  Edison Velasco-Sánchez, \\Francisco A. Candelas, and Fernando Torres
\thanks{Manuscript received: November, 29, 2023; Revised February, 22, 2024; Accepted April, 18, 2024.}
\thanks{This paper was recommended for publication by Editor A. Banerjee upon evaluation of the Associate Editor and Reviewers' comments.}
\thanks{This work has been supported by the regional Valencian Community Government and the European Union through the project PROMETEO/2021/075, as well as by the Spanish government through the grants PRE2019-088069 and PRE2022-101680 and the project PID2021-122685OB-I00.}
\thanks{Authors are with the Group of Automation, Robotics and Computer Vision (AUROVA), University of Alicante, San Vicente del Raspeig S/N, Alicante, Spain.
        {\tt\small miguelangel.munoz@ua.es}}%
\thanks{Digital Object Identifier (DOI): 10.1109/LRA.2024.3396055}
}

\begin{document}

\maketitle

\begin{abstract}
Feature-based geo-localization relies on associating features extracted from aerial imagery with those detected by the vehicle's sensors. This requires that the type of landmarks must be observable from both sources. This lack of variety of feature types generates poor representations that lead to outliers and deviations produced by ambiguities and lack of detections, respectively. To mitigate these drawbacks, in this paper, we present a dynamically weighted factor graph model for the vehicle's trajectory estimation. The weight adjustment in this implementation depends on 
 information quantification in the detections performed using a LiDAR sensor. Also, a prior (GNSS-based) error estimation is included in the model. Then, when the representation becomes ambiguous or sparse, the weights are dynamically adjusted to rely on the corrected prior trajectory, mitigating outliers and deviations in this way. We compare our method against state-of-the-art geo-localization ones in a challenging and ambiguous environment, where we also cause detection losses. We demonstrate mitigation of the mentioned drawbacks where the other methods fail.
 
\end{abstract}


\section{INTRODUCTION}
\label{sec:introduction}

\IEEEPARstart{A}{utonomous} navigation is a significant research topic because it can automate complex tasks using mobile robots, Unmanned Ground Vehicles (UGV), or self-driving cars. Navigating through an environment autonomously relies strongly on the localization module. The more extended approach for this purpose is \textit{Simultaneous Localization And Mapping} (SLAM) \cite{cadena2016past}, where the vehicle navigates building a model of the environment (the map) while simultaneously using it for self-localization. Alternatively, to simplify the localization, the mapping process could be avoided using an environment representation previously created by 
dedicated mapping vehicles~\cite{pauls2020monocular}. However, creating a map is usually expensive, especially for oversized areas. Moreover, it requires several loop closures for consistency, but despite this, a mapping process often accumulates minor errors that lead to global inconsistencies.

Over the last few years, the so-called geo-localization or geo-referencing has increased in importance in the literature. For the localization, this approach uses an environment representation obtained from aerial imagery. This avoids the expensive mapping process and the need for loop closures and provides implicit global consistency. We can distinguish two different strategies to perform geo-localization: \textit{end-to-end learned} \cite{tang2020rsl} and \textit{handcrafted-feature-based} \cite{munoz2022robust}.

The \textit{end-to-end learned} strategy uses the raw aerial image as an environmental representation while perceiving it with local sensors such as LiDAR \cite{li2023geo}, RADAR \cite{tang2020rsl}, or cameras~\cite{fervers2023uncertainty}. Then, it uses end-to-end learned models to extract dense features from both data sources to infer the vehicle's pose in the geo-referenced aerial image. In \cite{zhu2022transgeo,hu2020image,downes2022city}, the authors use a wide aerial image as a representation and infer the pose by crossing the sensor's information directly extracting the dense features from the whole image. In contrast, in \cite{li2023geo,fervers2022continuous,tang2023point}, the authors presented a pipeline that gets a crop around the prior pose to perform the end-to-end strategy.

The \textit{handcrafted-feature-based} geo-localization uses aerial imagery to extract handcrafted sparse features, while the same type of features should be detected from the sensors' data. Then, after data association, the vehicle's pose is estimated. This strategy implies a requirement: the type of feature used must be observable from both aerial and onboard vehicle sensors. In \cite{frosi2023osm,cho2022openstreetmap,kim2019fusing,roh2017aerial,yan2019global}, building walls are used as features, while in \cite{munoz2022robust,hu2019accurate,javanmardi2017towards}, the authors choose lane marking as landmarks that satisfy the observability requirement. In other works \cite{atia2019map,suger2017global,singh2020genetic}, the authors match the vehicle's trajectory with the lanes map, which is commonly named in the literature as map matching~\cite{huang2021survey}. However, we prefer to categorize it as a \textit{handcrafted} where the feature is the trajectory. The \textit{handcrafted-feature-based} strategy allows the measure of information in the detections before the pose inference. For example, in \cite{munoz2022robust}, the authors estimate prior confidence in the data and use it to self-tune the data association method depending on that confidence. For this work, we introduce the ground boundaries as a new feature type observable from both sources.

\begin{figure*}[t]
\centering
\includegraphics[width=500pt]{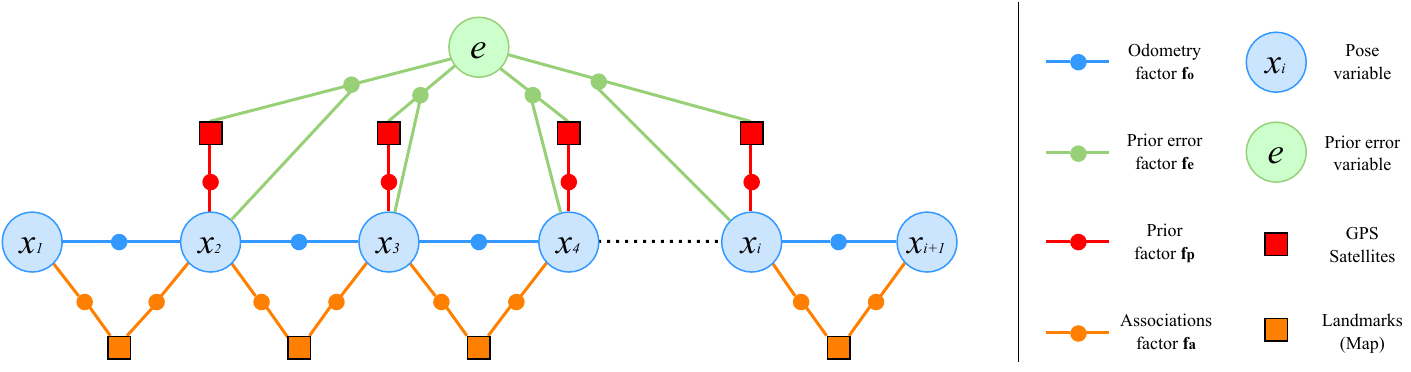}
\caption{\textbf{Proposed factor graph}: Each factor represents the difference between the observations and the model predictions. Section \ref{sec:factor_graph} explains the formalization of the (weighted) factors to obtain the residuals. The empty circles represent the variables that should be estimated by factor residual minimization. Finally, we denote by squares the elements that produce exteroceptive observations, i.e., GPS satellites and landmarks read from the map.}
\label{fig:factor_graph}
\end{figure*}

The mentioned observability requirement carries fewer features, leading to a sparse representation. This issue and the geo-referenced nature of the problem generate some drawbacks in the \textit{handcrafted-feature-based} strategy: (i)  A sparse map implies that some areas are ambiguous for the data association in the front direction of navigation, which can produce a considerable number of outliers. (ii) The poor variety of features introduces the risk of lack of detection in some navigation areas. (iii) Geo-localization needs a geo-referenced prior, such as GNSS (Global Navigation Satellite Systems). Those systems are usually precise but inaccurate, introducing offsets that vary smoothly through time and space, especially when there is a multipath problem \cite{del2020deeper}. In a previous work \cite{munoz2022robust}, we addressed the drawback (i). But with this approach, if we find a situation of type (ii), the localization converges to a prior trajectory that usually has an undesired offset, as we mentioned in (iii). Notably, the cited \textit{handcrafted-feature-based} geo-localization works usually don't pay special attention to avoid these drawbacks, so we consider it interesting to focus our research here.

This paper presents a \textit{handcrafted-feature-based} geo-localization that mitigates the exposed drawbacks through a dynamically weighted factor graph implementation in the vehicle's trajectory estimation. The weight adjustment in this model depends on information quantification in the detections performed using a LiDAR sensor. Furthermore, a prior (GNSS-based) error estimation is included in the model to avoid the drawback (iii). In this way, if, for example, we drive through an area ambiguous for data association, the weight is adjusted to trust more on the corrected prior trajectory, mitigating the drawback (i). In the case of the lack of detections (ii), again, the system will rely more on corrected prior.

In summary, our contributions are the following:

\begin{itemize}
    \item  A weighted factor graph that dynamically adjusts its weights depending on the information quantified from the detections. That produces a mitigation of the drawbacks (i), (ii), and (iii) previously mentioned.
    \item An information quantification strategy developed upon a previous one \cite{munoz2022robust,munoz2022lmr} that quantifies the information based on the associated map points instead of the raw detections. This quantification is the primary measure to adjust factor weights.
    \item A prior (GNSS-based) error estimation included in the model. With this corrected prior, we can hold the localization for low informative detections.
\end{itemize}


\section{WEIGHTED FACTOR-GRAPH}
\label{sec:factor_graph}

In this section, we formalize the proposed factor-graph model (Fig. \ref{fig:factor_graph}), where each factor is dynamically weighted depending on the information in the data (Section \ref{subsec:data_information}).  In this way, the more confident residuals will contribute mainly to the loss function in the optimization process, giving less importance to those who can generate unwanted minimums, for example, when outliers occur or when there is a lack of landmarks. 

The factors explained in Sections \ref{subsec:odom_factor}, \ref{subsec:prior_factor}, and \ref{subsec:asso_factor} are the most commonly implemented and the ones that generate the residuals to estimate the trajectory state variable $\mathbf{X} = ( \mathbf{x}_1, ... , \mathbf{x}_N )$, where each pose is $\mathbf{x}_i \doteq ( \mathbf{R}_i, \mathbf{t}_i)$, $\mathbf{t}_i \in \mathbb{R}^2$ is the translation, and $\mathbf{R}_i \in SO(2)$ is the rotation matrix. 

In contrast, in this work, we propose an additional factor explained in Section \ref{subsec:error_factor} that generates the residuals to estimate the error $\mathbf{e} = (e_x,  e_y)$ in the prior signal. The estimation of this variable allows corrections to the GPS observations, thus contributing to maintaining localization in outlier and lack regimes and avoiding undesired GPS errors in the loss function.

It is worth noting that this section explains the high-level formalization of the model (the \textit{back-end}). In contrast, in the next section (Section \ref{subsec:data_information}), we present the low-level (the \textit{front-end}) with more details, e.g., data association implementation, type of landmark, detections, etc.

\subsection{Odometry factor}
\label{subsec:odom_factor}
We assume we have an odometry system estimated from the LiDAR, cameras, IMU, and/or encoders. Then, given the relative transformations from consecutive frames $i$ and $i' = i - 1$ from the odometry trajectory $\hat{\mathbf{X}} = ( \hat{\mathbf{x}}_1, ... , \hat{\mathbf{x}}_N )$, and the poses in the estimated $\mathbf{X}$, we can define the odometry factors as follows:

\begin{equation}
    f^o_i = \omega^o_i \left\lVert \mathbf{R}^\mathsf{T}_i\left( \mathbf{t}_{i'} - \mathbf{t}_i \right) - \hat{\mathbf{t}}_{i,i'} \right\rVert^2_2 + \omega^o_i \left\lVert \mathbf{R}^\mathsf{T}_i\mathbf{R}_{i'} - \hat{\mathbf{R}}_{i,i'} \right\rVert^2_{F}.
    \label{eq:factor_odom}
\end{equation}

As we can see in \eqref{eq:factor_odom}, each norm is weighed by $\omega^o_i$. The value of that weight is dynamically obtained in each iteration depending on the quantification of the data information explained in Section \ref{sec:weight_adjustment}. When the data is considered poorly informative, this weight will acquire strength. The subscript $F$ in the second term indicates the Frobenius norm.

\subsection{Prior factor}
\label{subsec:prior_factor}
For geo-localization, it is essential to have a geo-referenced prior localization based on GNSS, which, in our case, is GPS. Such systems usually provide position information $\mathbf{t}^{g}_j = (x^g_j, y^g_j)$ as $j$-th observation. Given the time stamp availability, it is easy to obtain the association between GPS $j$-th observation and $i$-th odometry estimation. Then, we define the prior residuals as:

\begin{equation}
    f^p_j = \omega^p_j \left\lVert \mathbf{t}_i - \left( \mathbf{t}^{g}_j - \mathbf{\Bar{e}} \right) \right\rVert^2_2,
    \label{eq:factor_prior}
\end{equation}
where $\mathbf{\Bar{e}}$ is the estimated error that corrects the GPS observation, and the weight $\omega^p_j$ is calculated similarly to $\omega^o_i$ (Section \ref{sec:weight_adjustment}). Note that the bar in $\mathbf{\Bar{e}}$ indicates that, in this case, $\mathbf{e}$ is considered an observation, not a variable. With odometry and prior residuals, obtaining a so-called prior trajectory is possible. This trajectory provides satisfactory results in differential terms. But, due to the inaccuracy problems of GPS systems, the path usually has an offset that varies smoothly over time and space. We aim to correct this with the $\mathbf{e}$ estimation. However, to estimate $\mathbf{e}$, it is necessary to use a geo-referenced map to associate its information with local detections, obtaining a trusted localization, as explained in the following subsection. After fine $\mathbf{e}$ estimation, it is reasonable that when the data is considered poorly informative, $\omega^p_j$ and $\omega^o_i$ will acquire strength, and the final estimation will maintain their localization trusting on that corrected prior trajectory.

\subsection{Data associations factor}
\label{subsec:asso_factor}
As mentioned above, we have a geo-referenced world’s representation, defined as a set of landmark $\mathcal{L}$. Then, for each $i$-th frame in $\hat{\mathbf{X}}$, we observe the landmarks of the environment using the vehicle's sensors. From now on, we name these observations as detections $\mathcal{D}_i$. Using $\mathcal{L}$ and $\mathcal{D}_i$, we must perform a data association process, where its result is a set of pairs $((\mathbf{d}_{i_1}, \mathbf{l}_{i_1}), ... , (\mathbf{d}_{i_K}, \mathbf{l}_{i_K}))$. Given these associations, we can define the residuals between landmarks and detections as follows:

\begin{equation}
    f^a_i = \omega^a_i \sum_{k=1}^{K} \left\lVert \left(\mathbf{R}_i\mathbf{d}_{i_k} + \mathbf{t}_i \right) - \mathbf{l}_{i_k} \right\rVert^2_2.
    \label{eq:factor_asso}
\end{equation}

As shown in \eqref{eq:factor_asso}, the residual depends on the pose that transforms the detection from the local sensor to the map coordinates frame. In this case, in contrast to the odometry and prior residuals, the weight $\omega^a_i$ is strongest when the data is more informative.

\subsection{Prior error factor}
\label{subsec:error_factor}
In Section \ref{subsec:prior_factor}, we mentioned that the prior trajectory presents variable offset produced by GPS inaccuracies. Then, when we measure less informative data, and consequently, the prior path strengthens the optimization, the final localization could carry the mentioned inaccuracies. To avoid this effect, we estimate the prior error $\mathbf{e}$ to correct the GPS observation in \eqref{eq:factor_prior}. The factor for the error estimation is the following:

\begin{equation}
    f^e_j = \sum_{j'= j - w}^{j' = j} \omega^e_{j'} \left\lVert \mathbf{e} - \left( \mathbf{t}^{g}_{j'} - \Bar{\mathbf{t}}_{i'} \right) \right\rVert^2_2.
    \label{eq:factor_error}
\end{equation}

GPS error varies smoothly over time, so we estimate that variable using factors from limited past poses, e.g., from $j' = j-w$ to $j' = j$, being $i'$ the $i$-th position associated with $j'$. The notation of $\Bar{\mathbf{t}}_{i'}$ means that is the position of the state estimated, but in this case, it is used as observation instead of as a variable.

\subsection{Optimization}
The sum of all exposed factors is the cost function. Thus, the optimal state $\mathbf{X}^*,\mathbf{e}^*$ is such that it minimizes the said cost:

\begin{equation}
    \mathbf{X}^*,\mathbf{e}^* = \argmin_{\mathbf{X},\mathbf{e}} \left( \sum^N_i \left( f^o_i + f^a_i \right) + \sum^M_{j} \left( f^p_j + f^e_{j} \right) \right),
    \label{eq:optimization}
\end{equation}
where $M$ is the number of GPS observations, and $N$ is the number of odometry observations (that coincide with estimations). The weights explained in this section directly affect the cost function form, allowing us to avoid the problems exposed in Section \ref{sec:introduction}: (i) when the data information is insufficient with ambiguities risk, the system can prevent outliers by holding the trajectory taking strength on the corrected prior trajectory. (ii) the same occurs when we have a lack of detections. (iii) we avoid little deviations in the final estimation when correcting GPS inaccuracies.

The dynamic adjustment of these weights depends directly on the information in the data. The following section describes the information calculation and the consequent weight adjustment.


\section{DYNAMIC WEIGHT ADJUSTMENT}
\label{sec:weight_adjustment}

While in the previous section, we focused our explanation on the factor graph model, this section aims to expose the data information quantification and the adjustment of the weights for that model. In Section \ref{subsec:data_information}, we describe how to obtain the data information quantification. For that, we need to talk previously about landmarks, detection obtention, and the data association process. In section \ref{subsec:weights}, we specify how to adjust the weights as a function of the data information.

\subsection{Data information quantification}
\label{subsec:data_information}
In geo-localization, the type of landmark chosen for localization must be observable from aerial imagery and local sensors. The literature usually includes lane markings, vertical structures such as building walls, and even the vehicle's trajectory. In this work, we introduce another feature that satisfies the mentioned requirement: ground boundaries. This type of landmark is suitable for roads, city streets, and pedestrian areas such as university campuses.

In the following points, we describe how to obtain those detections $\mathcal{D}_i$ from a LiDAR sensor, get the map as a set of landmarks $\mathcal{L}$, and quantify the information after the data association.

\subsubsection{Detections}
\begin{figure}[t]
\centering
a) Front-view from LiDAR
\includegraphics[width=246pt]{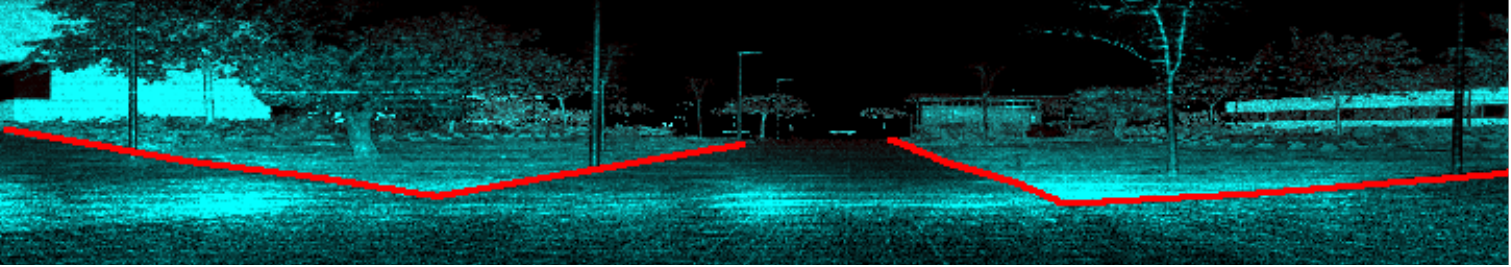}
\\b) Front-view mask for training
\includegraphics[width=246pt]{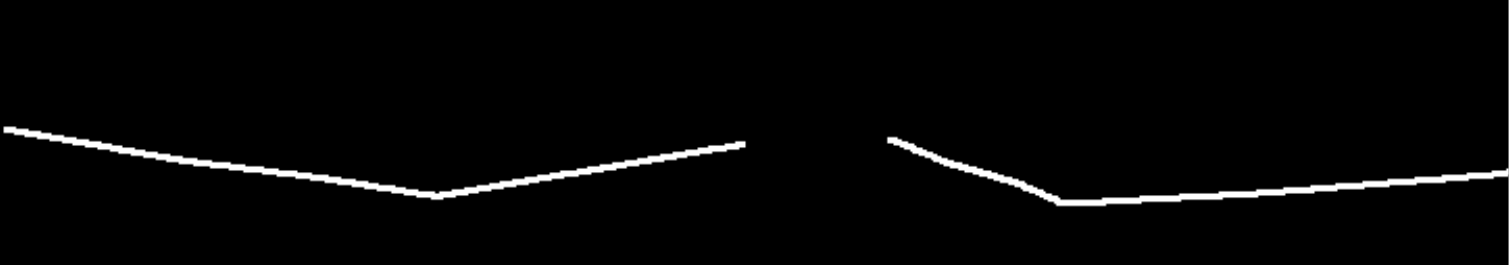}
\\c) Top-view from aerial image
\includegraphics[width=246pt]{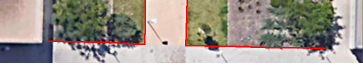}
\caption{\textbf{Example of $\mathcal{D}_i$ and $\mathcal{L}_i$}: a) Front-view from LiDAR information where the red line marks the ground boundaries ($\mathcal{D}_i$). b) Front-view mask extracted from (a). c) Top-view from the aerial image where the red line represents the exact boundaries shown in (a) and (b) ($\mathcal{L}_i$).}
\label{fig:landmarks}
\end{figure}

A LiDAR sensor usually provides a point cloud with 3D environment information and the reflectivity for each point. With this information, we mount a front-view representation in RGB image format. In this image, the GB channels contain reflectivity information, while in the R channel, we include the range image. In Fig. \ref{fig:landmarks} a), we show an example of a LiDAR scan as a front-view representation.

Then, we use the Convolutional Neural Network (CNN) Unet++ \cite{zhou2019unet++} with backbone resnet18 \cite{he2016deep} for ground boundary detections. To train the model, we generated our own dataset of 825 images and labeled them by hand, obtaining the ground truth masks (Fig. \ref{fig:landmarks} b)). We divided our dataset into $80\%$ for training and $20\%$ for testing with non-overlapping. Regarding generalization, the mentioned dataset was recorded in the Scientific Park of the University of Alicante, and we observed satisfactory results on the university campus, which is a different environment with more vegetation and a different pavement. In contrast, we needed to label new images in our experiments in KITTI, where the environment is totally different\footnote{The generalization of the complete localization method depends on that detection module but is independent of the contribution of this work. Roughly speaking, the detection module is like a black box in our approach. If we require more generalization, we would need to research a more sophisticated black box, which is out of the scope of this paper.}. After training, during the detection, each $u$-th pixel with value $1$ has its correspondence as a 3D point projected in 2D, being a detection $\mathbf{d}_{i_u} \in \mathcal{D}_i$, where $\mathbf{d}_{i_u} = (x^d_{i_u}, y^d_{i_u})$.

In a previous work \cite{munoz2022robust}, we used the polylines that describe detections to quantify the information using the angle between adjacent segments in the polyline. In this case, we observe that the boundaries in detections $\mathcal{D}_i$ are not arranged sequentially. We could process the data to obtain the polylines, which implies an undesired complex process in computational time terms. Thus, we opted to quantify the data using the structuration of the landmarks $\mathcal{L}$ associated with detections instead of directly using detections.

\subsubsection{Landmarks}
The landmarks $\mathcal{L}$ that form the map could also be detected with neural networks. Still, we use a handcrafted map generated by applications such as OpenStreetMaps \cite{haklay2008openstreetmap} to avoid post-processing. This implies that the map comprises a polyline set arranged in a friendly way to quantify the data using differential angles between adjacent segments in the polylines \cite{munoz2022lmr}. In Fig. \ref{fig:landmarks} c), we show labeled ground boundaries in an aerial image. These boundaries are the same labeled in front-view representations (Fig. \ref{fig:landmarks} a),b)).

Before $i$-th data association, we have detections set $\mathcal{D}_i$; then we must crop $\mathcal{L}$ around $\mathbf{x}_i$, obtaining $\mathcal{L}_i$. The $v$-th landmark $\mathbf{l}_{i_v} \in \mathcal{L}_i$ is defined as $\mathbf{l}_{i_v} = (x^l_{i_v}, y^l_{i_v}, \alpha_{i_v})$, where $\alpha_{i_v}$ is the differential angle between adjacent polyline segments.

Quantifying the information using $\mathcal{L}_i$ is unsuitable because it could contain landmarks not observed by the sensors, adding undesired information to quantification. In the next point, we describe quantifying the information after the data association process.

\subsubsection{Raw quantification from data association}
Given $\mathcal{D}_i$ and $\mathcal{L}_i$, we perform the data association process. In this work, we use an ICP (\textit{Iterative Closest Point}) due to its efficient implementation in PCL (\textit{Point Cloud Library}). First, we co-register  $\mathcal{D}_i$ with  $\mathcal{L}_i$, obtaining a new transformed set $\mathcal{D}'_i$. Finally, we find the closest point in $\mathcal{L}_i$ below a certain distance threshold for each $\mathbf{d}'_{i_u} \in \mathcal{D}'_i$. This process generates a set of associated pairs $((\mathbf{d}_{i_1}, \mathbf{l}_{i_1}), ... , (\mathbf{d}_{i_K}, \mathbf{l}_{i_K}))$ used in \eqref{eq:factor_asso}. Then, to quantify the information of the data rawly, we sum the values of the delta angle in the associated landmarks as:

\begin{equation}
    s_i = \sum^K_{k = 1} \alpha_{i_k}.
    \label{eq:information}
\end{equation}
This raw quantification is used to adjust the weights, as explained in the next section.

It is worth noting that this quantification based on polyline map representation is suitable for the rest of the feature types used in the literature, i.e., lane markings, building walls, or trajectories.

\subsection{Weights as a function of data information}
\label{subsec:weights}
As shown in \eqref{eq:information}, the raw information quantification is an accumulative value, where the minimum is $s^{min}_i = 0$, and its maximum depends on the environment, where our experiments observe a maximum $s^{max}_i \approx 60$. To obtain the $s^{max}_i$ value in a different area, we must drive only the part with more landmark density and calculate the maximum $s$ value. This result is intractable to weight the factor graph as we proposed in Section \ref{sec:factor_graph}. For this reason, to adjust the data association weight, we use a sigmoid function to restrict the quantification in a range $\left[0; 1\right]$:

\begin{equation}
    \omega^a_i(s_i) = \omega^e_j(s_i) = \frac{1}{1 + e^{-\Phi_i(s_i)}}.
    \label{eq:w_ae}
\end{equation}

As we can see in \eqref{eq:w_ae}, given $j$ and $i$ synchronization, we can adjust the prior error weight $\omega^e_j(s_i)$ as that $\omega^a_i(s_i)$. The form of the sigmoid depends on the function $\Phi_i(s_i)$, and we propose two different definitions compared in the evaluation section.

\begin{figure}[t]
\centering
\includegraphics[width=230pt]{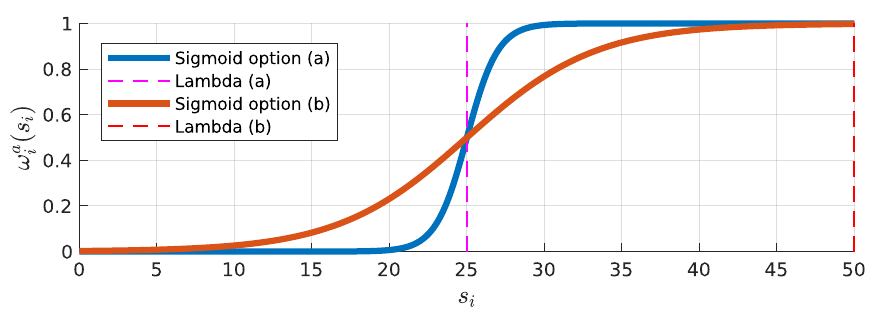}
\caption{Example of $\Phi^{(a)}_i(s_i)$ and $\Phi^{(b)}_i(s_i)$ application in a sigmoid function. The dotted lines indicate the values for $\lambda^{(a)}$ and $\lambda^{(b)}$.}
\label{fig:sigmoids}
\end{figure}

First, as an option (a), if we consider the range of information quantification as $s_i = \left[0; s^{max}_i\right]$ and if we take into account that a sigmoid function changes around zero in the range of $\left[-6; 6\right]$, we can displace the zero in $s_i$ so that $\Phi_i(s_i)$ remains as:

\begin{equation}
    \Phi^{(a)}_i(s_i) = \sum^K_{k = 1} \alpha_{i_k} - \lambda^{(a)},
    \label{eq:sigma_a}
\end{equation}
where $\lambda^{(a)}$ is a configurable parameter. In Fig. \ref{fig:sigmoids}, we show an example of this sigmoid configuration, where the pink dotted line marks the value of $\lambda^{(a)}$. This configuration provides a function that we can see as a smoothed step function where the parameter $\lambda^{(a)}$ tunes how restrictive the system is against information in the data. 

Second, as an option (b), we propose a smoothest function. In this case, we transform the sigmoid range $\left[-6; 6\right]$, where its size is $h$, to an information quantification range $\left[0; \lambda^{(b)}\right]$, where $\lambda^{(b)}$ is close to $s^{max}_i$:

\begin{equation}
    \Phi^{(b)}_i(s_i) = \frac{h}{\lambda^{(b)}} \sum^K_{k = 1} \alpha_{i_k} - \frac{h}{2}.
    \label{eq:sigma_b}
\end{equation}

In Fig. \ref{fig:sigmoids}, we show an example of this second option, where the red dotted line marks the value of $\lambda^{(b)}$.

Finally, we define the expressions to adjust the prior trajectory weights. First, for the odometry weight as:

\begin{equation}
    \omega^o_i(s_i) = \left(K_i + 1 \right) \left(2 - \omega^a_i(s_i)\right).
    \label{eq:w_op}
\end{equation}

And second, its variant for the prior as:

\begin{equation}
    \omega^p_j(s_i) = \frac{\left(K_i + 1 \right) \left(2 - \omega^a_i(s_i) \right)}{\left(\sigma^{x,y}_j + 1 \right)},
    \label{eq:w_p}
\end{equation}
where $K_i$ is the number of associations, this first term scales the prior trajectory weights to provide the same strength as the associations' residuals. We can see in \eqref{eq:w_op} that the second term provides strength when the data information is poor. In \eqref{eq:w_p}, $\sigma^{x,y}_j$ is the variance in $x,y$ plane for the GPS observations. Thus, noisily observations reduce the strength of the weight. The $+1$ regularizations in the three terms are to avoid zeros.


\section{EVALUATION}
\label{sec:evaluation}

\begin{figure}[t]
\centering
\includegraphics[width=190pt]{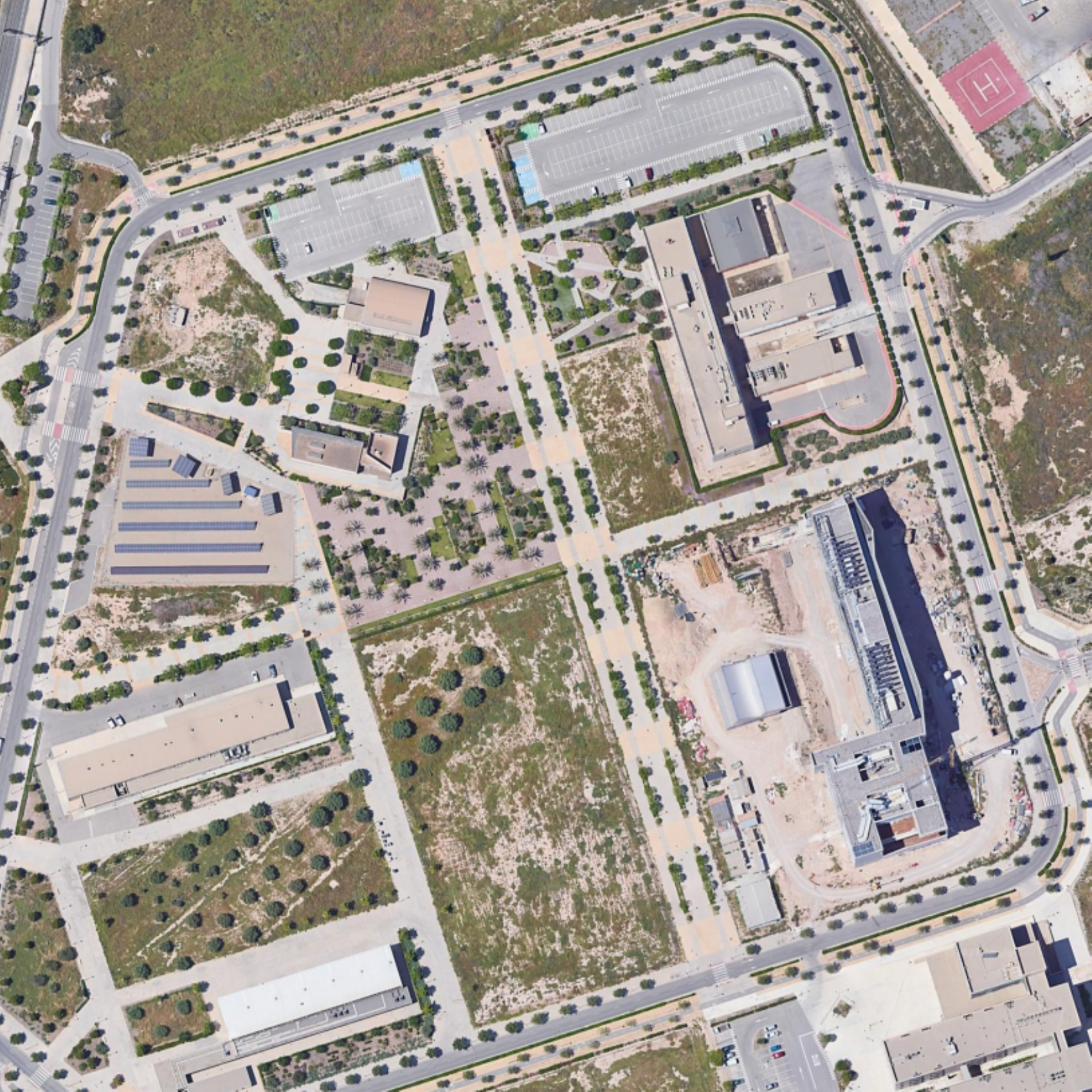}
\caption{Aerial image of the University of Alicante Scientific Park, where the evaluation was performed through circuits in Fig. \ref{fig:circuits}.}
\label{fig:aerial}
\end{figure}

\begin{figure}[t]
\centering
\includegraphics[width=236pt]{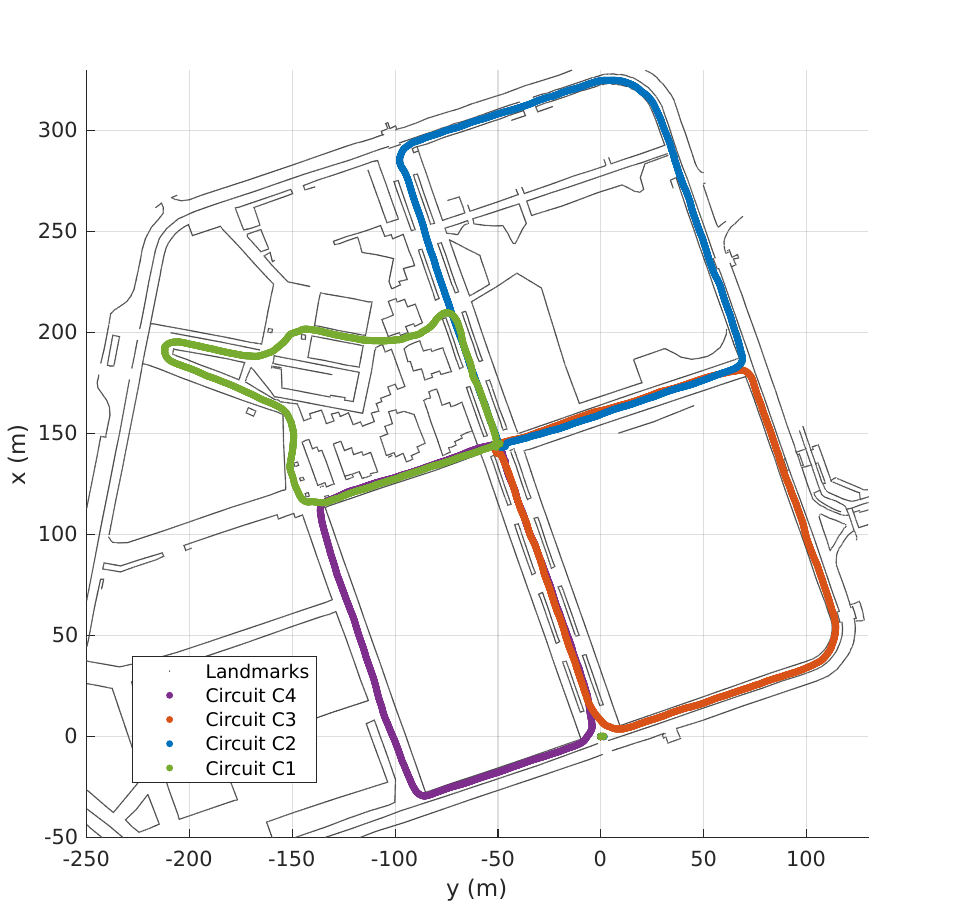}
\caption{The ground truth of the four circuits drove around the UA Scientific Park. The landmarks were obtained from the aerial image in Fig. \ref{fig:aerial}.}
\label{fig:circuits}
\end{figure}

\begin{figure*}[t]
\centering
\includegraphics[width=470pt]{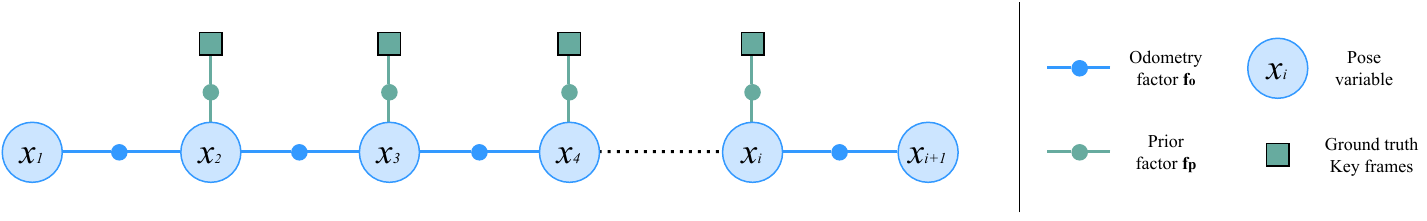}
\caption{\textbf{Factor graph for ground truth generation}: We use low-bias odometry as in Fig. \ref{fig:factor_graph}, while we use handcrafted positions as trusted keyframes generating residuals by prior factors. We obtain a whole ground truth trajectory by optimizing this model in offline mode.}
\label{fig:ground_truth}
\end{figure*}

\begin{table*}[t]
    \centering
    \caption{State-of-the-arts methods compared for the evaluation.}
    \begin{tabular}{c | c c c c c c}
          
         \textbf{Methods} & \textbf{Data source} & \textbf{Map source} & \textbf{Feature type} & \textbf{Model used}  & \textbf{Evaluated in} & \textbf{GPS}\\
    \hline
         Frosi \textit{et al.} \cite{frosi2023osm} & LiDAR & OpenStreetMap & Buildings & Factor graph & KITTI & No \\
         Own implementation \cite{frosi2023osm} & LiDAR & OpenStreetMap & Ground boundaries & Factor graph & Own & Yes \\
    \hline
         Cho \textit{et al.} \cite{cho2022openstreetmap} & LiDAR & OpenStreetMap & Buildings & None & KITTI & No \\
         Own implementation \cite{cho2022openstreetmap} & LiDAR & OpenStreetMap & Ground boundaries & Pose graph & Own & Yes \\
    \hline
         Muñoz-Bañón \textit{et al.} \cite{munoz2022robust} & LiDAR and Cameras & OpenStreetMap & Lane markings & Factor graph & Their Own & Yes  \\
         Own implementation \cite{munoz2022robust} & LiDAR & OpenStreetMap & Ground boundaries & Factor graph & Own & Yes \\
    \hline
         \textbf{Ours} & LiDAR & OpenStreetMap & Ground boundaries & Factor graph & KITTI/Own & Yes \\
    \hline
    \end{tabular}
    \label{tab:sota_methods}
\end{table*}

\begin{table*}[t]
    \centering
    \caption{Whole trajectories evaluation by Absolute Trajectory Error (ATE) in translation and rotation.}
    \begin{tabular}{c c|c c c c c c c c c}
          &  & \textbf{Ours} & \textbf{Ours} & \textbf{Ours} & \textbf{Ours} &  & \textbf{Muñoz-Bañón} & \textbf{Frosi} & \textbf{Cho}  \\
          
         \textbf{Session} & \textbf{ATE} & $\Phi^{(a)}$ \textbf{+} $\mathbf{e}$ & $\Phi^{(a)}$ \textbf{-} $\mathbf{e}$ & $\Phi^{(b)}$ \textbf{+} $\mathbf{e}$ & $\Phi^{(b)}$ \textbf{-} $\mathbf{e}$ & \textbf{Prior} & \textit{et al.} \cite{munoz2022robust} & \textit{et al.} \cite{frosi2023osm} & \textit{et al.} \cite{cho2022openstreetmap} \\
    \hline
         C1 & \textit{trans.} ($m$) & 0.147 & 0.493 & 0.165 & 0.552 & 2.032 & \textbf{\textcolor{blue}{0.098}} & 0.280 & 0.416 \\
           & \textit{rot.} ($\deg$) & 1.138 & 1.415 & 1.144 & 1.372 & 1.592 & \textbf{\textcolor{blue}{1.002}} & 1.358 & 1.453 \\
    \hline
         C2 & \textit{trans.} ($m$) & \textbf{\textcolor{blue}{0.225}} & 1.102 & 0.272 & 1.167 & 1.790 & 0.312 & 8.967 & - \\
           & \textit{rot.} ($\deg$) & 0.876 & 1.053 & \textbf{\textcolor{blue}{0.858}} & 1.140 & 0.916 & 0.989 & 4.896 & - \\
    \hline
         C3 & \textit{trans.} ($m$) & \textbf{\textcolor{blue}{0.113}} & 0.904 & 0.184 & 0.933 & 1.211 & 0.152 & 0.949 & 1.244 \\
           & \textit{rot.} ($\deg$) & 0.789 & 0.935 & 0.829 & 0.894 & \textbf{\textcolor{blue}{0.734}} & 0.887 & 1.107 & 1.766 \\
    \hline
         C4 & \textit{trans.} ($m$) & \textbf{\textcolor{blue}{0.096}} & 0.604 & 0.107 & 0.689 & 1.304 & 0.113 & 0.748 & 0.690 \\
           & \textit{rot.} ($\deg$) & \textbf{\textcolor{blue}{0.644}} & 0.725 & 0.653 & 0.744 & 0.764 & 0.892 & 0.885 & 0.856 \\
    \hline
         C1' & \textit{trans.} ($m$) & \textbf{\textcolor{blue}{0.230}} & 0.726 & 0.255 & 0.806 & - & 0.671 & 0.798 & 0.757 \\
           & \textit{rot.} ($\deg$) & \textbf{\textcolor{blue}{1.033}} & 1.772 & 1.202 & 1.679 & - & 1.283 & 1.750 & 1.913 \\
    \hline
         C4' & \textit{trans.} ($m$) & \textbf{\textcolor{blue}{0.113}} & 0.994 & 0.288 & 0.774 & - & 0.774 & 1.257 & 1.115 \\
           & \textit{rot.} ($\deg$) & \textbf{\textcolor{blue}{0.895}} & 1.140 & 0.937 & 1.137 & - & 1.104 & 0.996 & 1.093 \\
    \hline
    \end{tabular}
    \label{tab:rpe_results}
\end{table*}

This article argues that our contributions mitigate some undesired effects in geo-localization approaches. To demonstrate it, we focus our evaluation on that way. Before that, we evaluate whole trajectories in a general way, comparing different configurations of our method and three state-of-the-art \textit{handcrafted-feature-based} methods (Section \ref{subsec:whole_tr}). Due to our approach being in this category, we consider it as the best baseline for comparison. From there, we discuss the mitigation of the mentioned problems: (i) Section \ref{subsec:outlier_mitigation} shows how our approach mitigates the outliers produced by ambiguities. (ii) Section \ref{subsec:lm_mitigation} evaluates the mitigation of lack of detections. (iii) Section \ref{subsec:gps_error} discusses how our GPS error estimation can improve the results in a whole trajectory. Finally, we provide a comparison in KITTI for two methods \textit{end-to-end learned}.

\subsection{Setup}
The evaluation was performed in the University of Alicante (UA) Scientific Park (Fig. \ref{fig:aerial}). This pedestrian area is where we drove through the four circuits shown in Fig. \ref{fig:circuits}: C1, C2, C3, and C4. These circuits present some areas where the data is ambiguous for the data association being areas with outliers risk, especially C2, and C3. These parts are then adequate to evaluate (i). Moreover, to assess the effect (ii), we repeat two trajectories but eliminate the detections in some navigation parts. We name these repetitions C1' and C4'. Then, we consider that we have six paths for evaluation.

Regarding the mentioned ambiguity risk, we observe different challenge levels in these circuits. C1 has less risk due to passing areas with more corners, as we can see in Fig. \ref{fig:circuits}. C4 presents some straight regions, but others are informative. Finally, we consider C2 and C3 the more challenging because passes through large straight areas, especially C2.

We drove these circuits using our own developed UGV platform BLUE (\textit{roBot for Localization in Unstructured Environments}) \cite{del2020deeper,munoz2022openstreetmap,castano2023manipulacion}, which mounts a LiDAR Ouster OS1-128 for environment perception.

To obtain ground truth, we manually align the detections for some trajectory frames with the map, producing, in this way, some ground truth poses. We name these corrected poses as ground truth keyframes. Then, as we have low-bias LiDAR odometry \cite{velasco2023lilo}, we use it to interpolate the ground truth poses to whole trajectories. We perform this interpolation by optimizing the entire circuit using the odometry factors defined in \eqref{eq:factor_odom} and the prior factors described in \eqref{eq:factor_prior}, but in the last case, using the ground truth positions. In Fig. \ref{fig:ground_truth}, we show the graph model for the ground truth generation.

In Table \ref{tab:sota_methods}, we show the specifications of the state-of-the-art methods used for comparison. It is worth noting that, for a fair comparison, we implemented in C++ the methods cited as described in their papers but adapted to our implementation. e.g., by using GPS and ground boundary features. For \cite{frosi2023osm}, we included the GPS factor in their own factor graph model. In the case of \cite{cho2022openstreetmap}, we generated a pose graph including the poses calculated using the descriptor presented in \cite{cho2022openstreetmap}, the odometry, and the GPS factors. This GPS augmentation doesn't contradict the contributions of the papers because in \cite{frosi2023osm}, the authors comment that GPS is an "optional" signal in their method, while in \cite{cho2022openstreetmap}, the authors don't aim to replace GPS; they seek to replace LiDAR maps. Table \ref{tab:sota_methods} specifies the differences indicating our implementation. We use the Absolute Trajectory Error (ATE) metric for these comparisons. 

\begin{figure}[t]
\centering
\includegraphics[width=230pt]{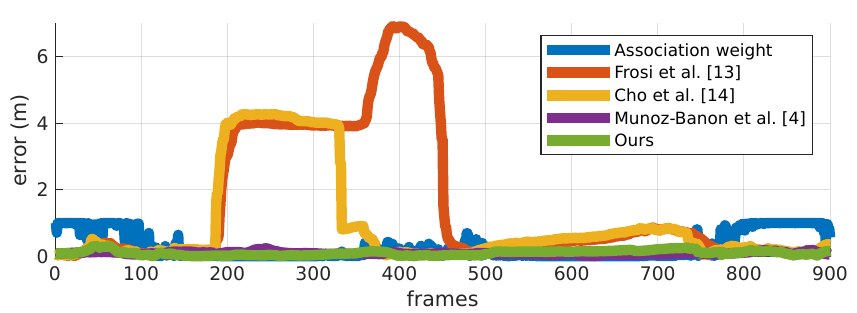}
\includegraphics[width=230pt]{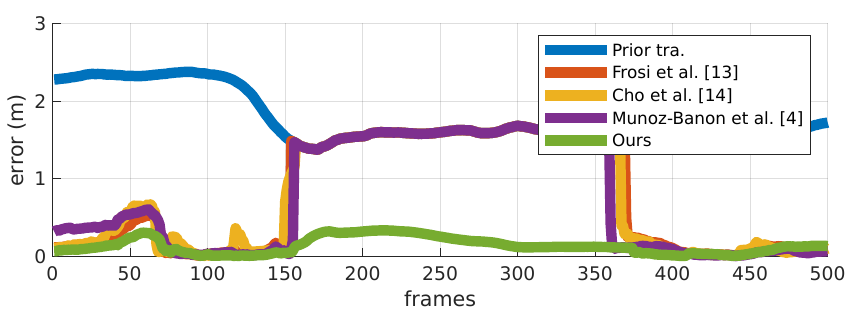}
\caption{A comparison of \textbf{ATE evolution per frame} for each compared method, where we evaluate the behavior for \textit{top: outlier mitigation} and for \textit{bottom: mitigation of detection losses}. }
\label{fig:outliers}
\end{figure}

\subsection{Whole trajectories evaluation}
\label{subsec:whole_tr}
Table \ref{tab:rpe_results} shows the results for the whole trajectory evaluation. The first four columns are the combinations of different configurations of our approach by combining the two proposed $\Phi$ functions with including or not the prior error estimation $\mathbf{e}$. We also show the prior trajectory and the three state-of-the-art methods compared results. The blue-marked values mean the best outcome for each circuit.

With the combination of our approach, we observe that the $\Phi^{(a)}$ + 
$\mathbf{e}$ implementation produces slightly better results than $\Phi^{(b)}$ + 
$\mathbf{e}$. In those cases, we mitigate the effect (i) even in the more challenging circuits C2 and C3. In both cases, we see that the error in C1' and C4' is held, mitigating the effect (ii). When we eliminate $\mathbf{e}$ estimation, the errors increase due to an ambiguous part of the circuit converging to a non-corrected prior trajectory. 

Regarding the method Muñoz-Bañón \textit{et al.} \cite{munoz2022robust}, we can see that the effect (i) is mitigated for the circuits C1 to C4. However, when detections are lost, the method converges to non-corrected prior, increasing the error by the effect of drawback (ii).

Methods Frosi \textit{et al.} \cite{frosi2023osm} and Cho \textit{et al.} \cite{cho2022openstreetmap} cannot mitigate either (i) and (ii), and the errors become too large in the most challenging circuits C2 and C3. Cho \textit{et al.} \cite{cho2022openstreetmap} did not finalize the most challenging circuit, C2, because it became lost. Method Frosi \textit{et al.} \cite{frosi2023osm} can complete that circuit but gets lost in some parts. Even when GPS augments \cite{cho2022openstreetmap,frosi2023osm}, they can be lost because, without weight adjustment, the LiDAR has more observations and produces more residuals in the optimization process. For the same reason, the ambiguity effect can produce results that overcome the prior error.

Looking into the results shown in Table \ref{tab:rpe_results}, we can infer how problems (i) and (ii) affect each method, but in the following section, we depict some concrete examples.

The Unet++ is a high-speed network, and we process images with low resolution (OS1-128 LiDAR resolution, 128x2048). Then, the experiments were performed in real-time, where our loop spent $57ms$ for the whole process, less than the $100ms$ required for real-time, which is the period of the LiDAR sensor. Muñoz-Bañon \textit{et al}. and Frosi \textit{et al}. spent around $90ms$, while Cho \textit{et al}. occupied $170ms$, which involves processing 1 of each two scans for real-time implementation. The experiments have been performed on an i7-7700HQ CPU with 16 GB of RAM in C++. The network was implemented in PyTorch using a GPU GTX 1050 Ti.

\subsection{Outlier mitigation}
\label{subsec:outlier_mitigation}
To evaluate outlier mitigation in more detail, we crop a trajectory through an area with outlier risk, i.e., a place where the $\omega^a_i$ has values close to zero. Then we evaluate the error against the ground truth per each frame for the compared methods. 

In Fig. \ref{fig:outliers} (\textit{top}), we can see the results of that process, where we show the value of $\omega^a_i$ in blue. When this value is near zero, the data is non-informative, and there is an outlier risk. The methods Frosi \textit{et al.} \cite{frosi2023osm}, and Cho \textit{et al.} \cite{cho2022openstreetmap} increase their errors, while Muñoz-Bañón \textit{et al.} \cite{munoz2022robust} and our proposed approach both mitigate the drawback (i). 

\subsection{Mitigation of detection losses}
\label{subsec:lm_mitigation}
As mentioned in the setup section, we caused a lack of detections in parts of circuits C1 and C4, being then these circuits as C1' and C4'. Then, to look into the trajectories in detail to evaluate the effect (ii) mitigation, as in the previous section, we crop a path through an area where we stopped the detection process. Finally, we evaluate the error against the ground truth per each frame for the compared methods.

In Fig. \ref{fig:outliers} (\textit{bottom}), we can see the results of that process, stopping the detections between the frames $150$ and $350$ approx. The prior trajectory error is shown in blue. We can see how when the detections are blocked, the errors in all compared methods become the same value as the prior error. In contrast, our approach can maintain stable error by mitigating the drawback (ii).

\subsection{Effects of GPS error estimation influences}
\label{subsec:gps_error}
As a drawback (iii), we argue that inaccuracies in the GNSS-based prior trajectory introduce errors in the final estimation through the prior factor \eqref{eq:factor_prior}. Evaluating how $\mathbf{e}$ estimation can improve the final pose inference in areas where there is no lack and no ambiguities is complicated because in our approach when there is no $\mathbf{e}$ estimation, the ATE increases because of ambiguities areas. Plot areas with no risk do not provide enough information. As a possible way to get some clue, we propose looking into circuit C1 results because it is the one where less ambiguities risk. In this case, we can see that our method is worst when $\mathbf{e}$ is not estimated.

\subsection{Comparison with end-to-end learned methods}
\label{subsec:end2end_comparison}

In the previous sections, we evaluate the mitigation of the typical drawbacks led by \textit{handcrafted-feature-based} methods. However, evaluating our approach compared with \textit{end-to-end learned} techniques is interesting. In this way, apart from demonstrating the mitigation of the discussed weaknesses, we show that our method is state-of-the-art for all geo-localization strategies. It is worth noting that the \cite{fervers2022continuous,li2023geo} approaches don't use GPS, and we don't augment it because in their methods, as in most \textit{end-to-end learned}, the authors present their strategies as GPS replacements.

We implemented our approach in the KITTI Odometry Benchmark using the provided odometry and LiDAR information (range and reflectivity). We labelled the road boundaries for $938$ images from the different Odometry Benchmark scenes. Then, we divided our dataset into $80\%$ for training and $20\%$ for testing with non-overlapping, obtaining $72\%$ of the IoU metric as individual frame performance. Table \ref{tab:rpe_results_kitti} shows the result for sequences 00, 07, 09, and 10. We chose the sequences to have diverse environments and roads. The empty values indicate that the author's paper doesn't provide results for such a sequence.

\begin{table}[ht]
    \centering
    \caption{Whole trajectories evaluation (ATE) in KITTI.}
    \begin{tabular}{c|c c c c}
         \textbf{Method} & \textbf{Sec. 00} & \textbf{Sec. 07} & \textbf{Sec. 09} & \textbf{Sec. 10} \\
    \hline
         Fervers \textit{et al}. \cite{fervers2022continuous} ($m$) & - & 0.85 & - & 0.96 \\
    \hline
         Li \textit{et al}. \cite{li2023geo} ($m$) & - & \textbf{\textcolor{blue}{0.44}} & 1.16 & 0.93 \\
    \hline
         Ours $\Phi^{(a)}$ \textbf{+} $\mathbf{e}$ ($m$) & \textbf{\textcolor{blue}{0.35}} & 0.47 & \textbf{\textcolor{blue}{0.22}} & \textbf{\textcolor{blue}{0.31}} \\
    \hline
    \end{tabular}
    \label{tab:rpe_results_kitti}
\end{table}

\section{CONCLUSIONS}
\label{sec:conclusions}

This paper presented a geo-localization approach based on a weighted factor graph that dynamically adjusts its values depending on the information measured in the data. Moreover, the GNSS-based prior error estimation is included in the model. This strategy mitigates typical drawbacks in the \textit{handcrafted-feature-based} geo-localization approaches: (i) The outlier raised from ambiguous representation. (ii) The deviations produced for sparse representations. (iii) The errors introduced by the GNSS-based prior. We demonstrate those mitigations experimentally by improving recent state-of-the-art methods in this way.


\bibliography{references.bib}{}
\bibliographystyle{IEEEtran}

\end{document}